\documentclass[letterpaper, preprint, paper,11pt]{AAS}	

\usepackage{bm}
\usepackage{amsmath}
\usepackage{amssymb}
\usepackage{subcaption}
\usepackage{soul}   
\usepackage[colorlinks=true, pdfstartview=FitV, linkcolor=black, citecolor= black, urlcolor= black]{hyperref}
\usepackage{overcite}
\usepackage{footnpag}			      	
\usepackage{xcolor}
\usepackage{ulem}   
\usepackage{multirow}

\PaperNumber{21-238}

\begin{document}

\title{Genetic Fuzzy System-Based Multi-robot Coordination for planetary missions}

\author{Daegyun Choi\thanks{PhD Student, Department of Aerospace Engineering and Engineering Mechanics, University of Cincinnati, Cincinnati, OH 45221, USA.}
\ and Donghoon Kim\thanks{Assistant Professor, Department of Aerospace Engineering and Engineering Mechanics, University of Cincinnati, Cincinnati, OH 45221, USA.}
}

\maketitle{}

\begin{abstract}
This paper proposes a decentralized approach for a multi-robot system (MRS) using a genetic fuzzy system to perform a collaborative object transportation task that minimizes the total path length of the MRS in unstructured environment while avoiding obstacles. For an environment given by an elevation map, terrain traversability analysis with respect to the slope is performed to reduce the dimension and identify non-traversable areas that can be considered as obstacles, and the given map is converted into a traversability map in two dimensional space. In the training process, proposed fuzzy inference systems (FISs) to generate the MRS's velocity for transporting an object to a target position are optimized by a genetic algorithm with several scenarios, such as a local minima, a target that is close to an obstacle, and a cluttered environment. The trained FIS models are applied to the testing environment, which is the converted traversability map, and validated using multiple scenarios.
\end{abstract}

\section{Introduction}

As increasing the interests in other planets, several space exploration missions including planetary missions are planned\cite{moon2020}. Among several planets within our solar system, Mars is the most attractive planet for human exploration because of the characteristics of Mars that provides an atmosphere, moderate temperatures, and the nearly identical day length, etc \cite{Mars2020,Sonsalla2017,Halbach2017}. Currently, robotic explorations for the geological and biological information of Mars and for preparing human missions are planned \cite{Mars2020,Sonsalla2017}. In particular, for human missions in Mars, it is required to construct an outpost and/or (re)locate infrastructures to perform a long-duration scientific expedition to an extreme environment. Those activities require robots with a transporting capability to carry an object. 
Up to now, several space explorations missions have been performed by a single robot system that has multiple functions and high-power capacity. Also, since a single robot system requires a complex mechanism with multiple sensors to support various missions, it requires high cost to build and manage and contains high risks of mission failure \cite{Yan2013,Huang2020}. 
On the other hand, using multiple and small robotic platforms operating cooperatively and collaboratively can overcome a single robot's limitations and reduce risks. A multi-robot system (MRS) requires a relatively lower cost and risks due to its simplicity. Furthermore, since it provides several advantages, such as greater redundancy, better system reliability, and flexibility, an MRS is appropriate for planetary missions, especially transporting a large and/or massive object in other planet.

With multiple benefits of an MRS, transporting an object using an MRS has been studied in recent decades. To tackle this problem, a multi-robot manipulation algorithm which allows the MRS to move an object along with a desired trajectory to a goal location is proposed \cite{Wang2015, Wang2016}. The robots coordinate their actions through sensing the motion of the object without an explicit communication network among robots, and a force consensus technique using the sensing information is applied to achieve the mission. In the study in Reference~\citenum{Chand2019}, a deformable object transportation problem is dealt with by using a leader-follower formation control algorithm. A path planning algorithm is used to avoid static obstacles for the virtual leader, and a constrained optimization method for multi-robot formation control is proposed. Also, in the research investigated by Alonso-Mora et al.\cite{Alonso-Mora2017}, a large obstacle-free convex region as a local planner is considered, and the parameters of the formation are optimized by sequential convex programming. Then, a global path planning is performed via the constrained optimization. Furthermore, various approaches for transporting an object have been proposed using approaches for the wavefront algorithm \cite{Fraile2008} and a fuzzy sliding mode control \cite{Dai2016}, etc. Plenty of existing collaborative control strategies use a centralized controller and consider even flat surfaces to perform missions. However, these control techniques may not be preferable for space exploration missions in rough terrain environment. Therefore, a decentralized approach for an MRS using fuzzy inference systems (FISs) trained by a genertic algorithm (GA) is proposed to perform a collaborative task, which is to transport an object to a designated area with the shortest path length in unstructured environments while avoiding obstacles.

\section{Environment and problem formulations}

Terrain information in other planets can be obtained by the digital elevation map (DEM) that is acquired by reconnaissance orbiters, such as Lunar Reconnaissance Orbiter or Mars Reconnaissance Orbiter. Since it is difficult to directly use the DEM for path planning, navigation, etc., terrain traversability analysis (TTA) is firstly performed with respect to the slope at each grid to reduce the space from three-dimension (3-D) to two-dimension (2-D). The role of TTA is to transform an elevation map into a traversability map based on the slope information at each position.

The DEM provides the elevation data as well as the information for the center location of the given map and a map scale as a distance per pixel. The center location is generally given by latitude and longitude angles, and $x$ and $y$ coordinates corresponding to those angles are computed by latitude/longitude projection.
Then, a square shaped terrain patch centered at $x_i$ and $y_j$ is defined as $P = \{ z_{kl}|k = i-L, \dots, i+L; \ l = j-L, \dots, j+L \}$ over all information, where $L$ is the positive integer. The number of data point in $P$ is $N = (2L+1)\times (2L+1)$, and the data in $P$ fits to a plane in the sense of least-square approach. To find the plane, the following matrix is constructed using the data in $P$ as follows\cite{Ye2007,Gu2008}:
\begin{equation}
    Q = \begin{bmatrix} x_1-\bar{x}& x_2-\bar{x} & \cdots & x_N-\bar{x} \\ 
    y_1-\bar{y}& y_2-\bar{y} & \cdots  & y_N-\bar{y} \\
    z_1-\bar{z}& z_2-\bar{z} & \cdots  & z_N-\bar{z}
    \end{bmatrix}
    \begin{bmatrix} x_1-\bar{x}& y_1-\bar{y} & z_1-\bar{z} \\ 
    x_2-\bar{x}& y_2-\bar{y} & z_2-\bar{z} \\
    \vdots & \vdots & \vdots \\
    x_N-\bar{x}& y_N-\bar{z} & z_N-\bar{z}
    \end{bmatrix}
\end{equation}
where $z_m$ is the elevation value of $m$-th data with coordinates $x_m$ and $y_m$ for $m=1,2, \dots, N$, and $\bar{x}$, $\bar{y}$, and $\bar{z}$ are defined as\cite{Ye2007,Gu2008}
\begin{equation}
    \bar{x} = \frac{1}{N}\sum_{m=1}^{N}x_m, \quad \bar{y} = \frac{1}{N}\sum_{m=1}^{N}y_m, \quad \bar{z} = \frac{1}{N}\sum_{m=1}^{N}z_m
\end{equation}

Then, the eigenvalues of $Q$ and their corresponding eigenvectors are computed, and the normal vector to the least-square plane is obtained as the eigenvector corresponding to the minimum eigenvalue. Thus, the slope of the plain obtained by $P$ is calculated by\cite{Ye2007,Gu2008}
\begin{equation}
    \gamma_{ij} = \cos^{-1}({\bf h}_{ij}\cdot{\bf n}_3)
\end{equation}
where ${\bf h}_{ij}$ is the normal vector to the least-square plane and ${\bf n}_3$ is the unit vector that is normal to the horizontal plane. When the slope at the grid is less than the threshold that is related to the robots' capability that maintains an object with stable attitude, this grid is assumed as a traversable area. Through this process, the elevation map is transformed into the traversability map, which is composed of traversable and non-traversable areas.


\begin{figure}[htbp]
    \centering
    \includegraphics[width=0.5\columnwidth]{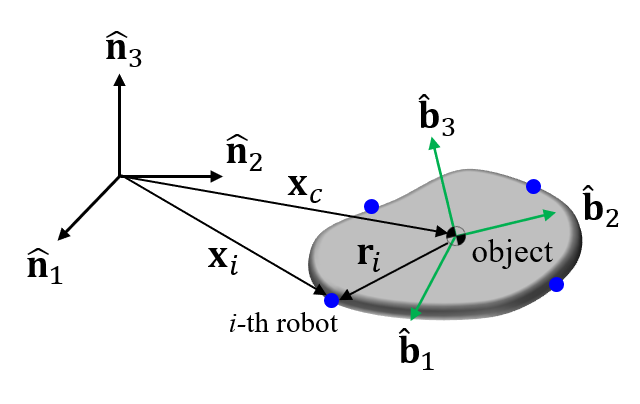}
    \caption{Definitions for frames and vectors}
    \label{fig:frame_def}
\end{figure}

To describe motions of robots and an object, frames and vectors are defined as shown in Figure \ref{fig:frame_def}. The inertial and body frames are defined using unit vectors $\hat{\bf n}_k$ and $\hat{\bf b}_k$ for $k$ = 1,2, and 3, respectively. The position and velocity vectors of $i$-th robot are defined as ${\bf x}_i \in \mathbb{R}^3$ and $\dot{\bf x}_i \in \mathbb{R}^3$, and the object's position and velocity vectors are defined as ${\bf x}_c \in \mathbb{R}^3$ and $\dot{\bf x}_c \in \mathbb{R}^3$. The attitude (yaw, pitch, and roll angles) of the object with respect to the inertial frame are defined as $\psi$, $\theta$, and $\phi$ with the 3-2-1 set of Euler angles, and its angular velocity is defined as $\bm\omega \in \mathbb{R}^3$. Given the transformed 2-D environment (or the traversality map) through the TTA, it is assumed that all robots have a capability that makes the object to be transported with stable attitude. That is, if no or very small rotational changes along roll and pitch axes can be maintained, it can be assumed that the object makes the planar motion. When each robot is assumed as a particle in 2-D space to simplify the problem, motions of the object and all robots are described in x-y plane made by $\hat{\bf n}_1$ and $\hat{\bf n}_2$.

With two components for the position and velocity vectors in x-y plane and one component for the attitude and angular velocity along with z-axis made by $\hat{\bf n}_3$, the changes of the velocity for each robot make the translational and rotational motion of the object. The kinematics between $i$-th robot and the object's center of mass are defined as
\begin{equation}\label{eq:kin}
    \dot{\bf x}_i = \dot{\bf x}_c + \bm\omega \times {\bf r}_i
\end{equation}
where $i=1,2,...,n$. Since the object is moved on the x-y plane, the unknown states for the object are $\dot{x}_{c,x}$, $\dot{x}_{c,y}$, and $\omega_{z}$. Rearranging Eq. \eqref{eq:kin} for all robots in terms of the unknown states yields the following:
\begin{equation}
    \begin{bmatrix}
        1 & 0 & -r_{1,y} \\ 0 & 1 & r_{1,x} \\ & \vdots & \\ 
        1 & 0 & -r_{n,y} \\ 0 & 1 & r_{n,x}
    \end{bmatrix}
    \begin{bmatrix}
        \dot{x}_{c,x} \\ \dot{x}_{c,y}  \\ \omega_z
    \end{bmatrix}
    =
    \begin{bmatrix}
        \dot{x}_{1,x} \\ \dot{x}_{1,y} \\ \vdots \\ \dot{x}_{n,x} \\ \dot{x}_{n,y}
    \end{bmatrix}
\end{equation}

Therefore, when each robot's velocity vector and its position vector with respect to the center of mass of the object are given, the object's states are obtained by using the least square method. Then, the position and orientation of the object are obtained by the sum of the position at the previous time step and the velocity multiplied by the time interval, and each robot's position information is calculated similarly.  In this research, the velocity of each robot is determined by FIS models based on the information about a target position and the nearest obstacle. Note that non-traversable areas in the traversability map are assumed as obstacles.

\section{Proposed genetic fuzzy system model}

\begin{figure}[htbp]
    \centering
    \begin{subfigure}[b]{0.45\columnwidth}
        \centering
        \includegraphics[width=0.7\columnwidth]{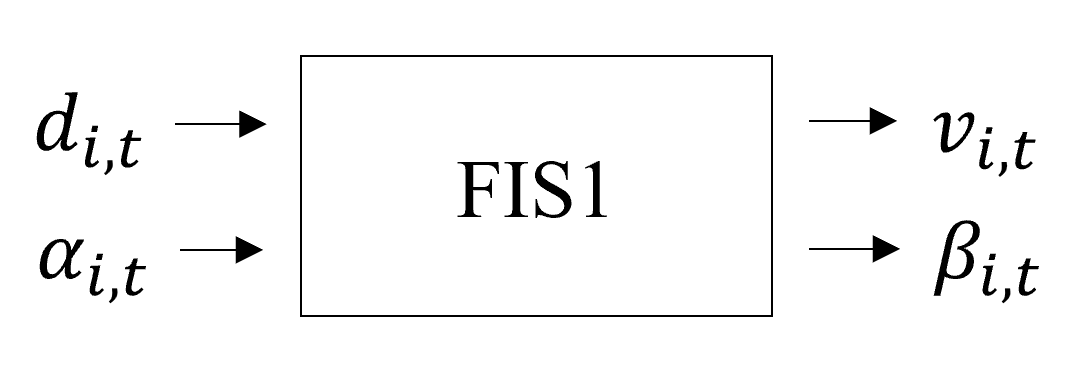}
    \end{subfigure}
    \begin{subfigure}[b]{0.45\columnwidth}
        \centering
        \includegraphics[width=0.7\columnwidth]{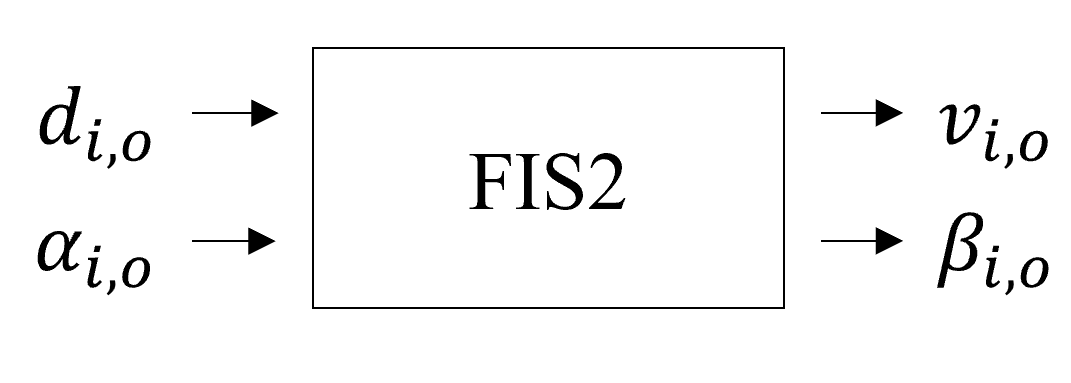}
    \end{subfigure}
    \caption{Proposed FIS models}
    \label{fig:fis_model}
\end{figure}

The robots' positions at every time step are determined by their velocities that are obtained by the FISs proposed. To compute the velocity for the robots, each robot uses two FISs that have two inputs and two outputs for each FIS as shown in Figure \ref{fig:fis_model}. Note that FIS1 determines a magnitude ($v_{i,t}$) and a correction angle ($\beta_{i,t}$) for $i$-th robot's velocity vector heading to the target position, and FIS2 determines a magnitude ($v_{i,o}$) and a correction angle ($\beta_{i,o}$) for $i$-th robot's velocity vector avoiding obstacles, respectively. 


FIS1 requires each distance $d_{i,t}$ between $i$-th robot and the target position and each angle $\alpha_{i,t}$ between $i$-th robot's velocity vector and the relative position vector ${\bf x}_{t/i}$ from $i$-th robot's position to the target position as inputs. Each input is computed as
\begin{equation}\label{eq:fis1_input}
    d_{i,t} = ||{\bf x}_{t/i}||, \quad \alpha_{i,t} = \cos^{-1}\left( \frac{\dot{\bf x}_i^T {\bf x}_{t/i}}{||\dot{\bf x}_i|| \ ||{\bf x}_{t/i}||} \right)
\end{equation}
where ${\bf x}_{t/i}$ is defined as ${\bf x}_t - {\bf x}_i$ and ${\bf x}_t$ is the target position vector. Also, the inputs of FIS2 are defined as each distance $d_{i,o}$ between $i$-th robot's position and the nearest obstacle's position and each angle $\alpha_{i,o}$ between $i$-th robot's velocity vector and the relative position vector ${\bf x}_{o/i}$ from $i$-th robot's position to the nearest obstacle's position. Both inputs for FIS2 are described as
\begin{equation}\label{eq:fis2_input}
    d_{i,o} = ||{\bf x}_{o/i}||, \quad \alpha_{i,o} = \cos^{-1}\left( \frac{\dot{\bf x}_i^T {\bf x}_{o/i}}{||\dot{\bf x}_i|| \ ||{\bf x}_{o/i}||} \right)
\end{equation}
where ${\bf x}_{o/i}$ is defined as ${\bf x}_o - {\bf x}_i$ and ${\bf x}_o$ is the position vector for the nearest obstacle. With those inputs calculated by Eqs. \eqref{eq:fis1_input} and \eqref{eq:fis2_input}, each FIS provides two outputs, and the velocity vectors for $i$-th robot are calculated by using the outputs as follows:
\begin{align}\label{eq:fis1_vel}
    {\bf v}_{i,t} & =  [v_{i,t} \cos(\beta_{i,t} +\rho_i), \ v_{i,t} \sin(\beta_{i,t} +\rho_i)]^T \\ \label{eq:fis2_vel}
    {\bf v}_{i,o} & =  [v_{i,o} \cos(\beta_{i,o} +\rho_i), \ v_{i,o} \sin(\beta_{i,o} +\rho_i)]^T 
\end{align}
where $\rho_i$ is $i$-th robot's heading angle that represents the direction of the velocity vector of the robot.Therefore, the velocity input for each robot is finally obtained by adding the two vectors as ${\bf v}_{i,t} + {\bf v}_{i,o}$. In fact, each robot only utilizes the information of the robot itself to determine the velocity input as a decentralized MRS.

\begin{figure}[htbp]
    \centering
    \includegraphics[width=0.9\columnwidth]{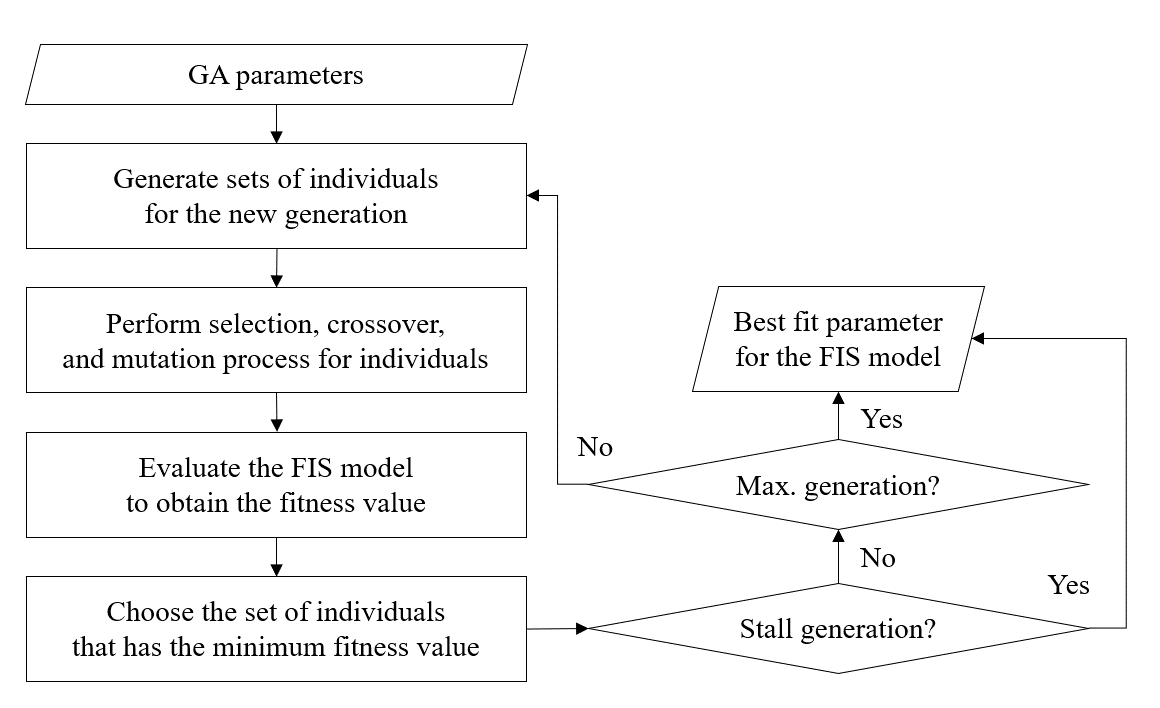}
    \caption{Flowchart for the training process}
    \label{fig:train_flow}
\end{figure}

For the two FISs, it considers the triangle shape of membership functions to simplify the problem and reduce the number of parameters, and some of the edges in the membership functions and the rules in the rulebase are optimized by the GA. The total number of parameters to be optimized is 20 for the membership functions and 54 for the rules in the rulebase. For the defuzzification process to obtain the output value, the centroid method that is the most common method is utilized. The training process is displayed in Figure \ref{fig:train_flow}. 

At the beginning of the training process, it requires the initialization for the GA parameters that include the number of generations, population size, and the stall number of generation, etc. Then, sets of individuals, which represent FIS parameters to be optimized, are randomly generated up to the population size. After that, the selection, crossover, and mutation process, which are the main contribution of the GA, are applied to sets of individuals. Once this process is completed, the evaluation of the fitness function is performed using the FIS models. Note that each set of individuals is substituted into the FIS parameters for the evaluation. Among several sets of individuals, the set of individuals that has the minimum fitness value is chosen as the best fit solution at the current generation, and it checks the convergence criteria that are defined as the stall number of generations and the maximum number of generations. If the fitness value is not changed during the stall number of generations or the maximum number of generations is achieved, the training process is terminated, and the best fit solution that has the minimum fitness value is selected for the FIS models. Note that the evaluation function for the optimization, which is known as the fitness function for the GA, is defined as
\begin{equation}\label{eq:fit}
    f_{\text{fit}} =  \sum_{i=1}^{n}d_i + \zeta
\end{equation}
where $d_i$ is the total path length for $i$-th robot to reach the target position, and $\zeta$ is a penalty that is applied for any undesirable path. The penalty is defined as a collision between each robot and the obstacles and a situation where any of the robots is out of the tolerance range at the end of the simulation.


\section{Simulation studies}

\subsection{Descriptions of training and testing environments}

In the field of collision avoidance (CA), there exist two well-known situations that affect CA performance. A situation is called local minima where an obstacle locates exactly between the agent and the target position, and there is a high possibility that the agent gets stuck in front of the obstacle. Also, if an obstacle locates near the target position, the agent cannot reach the target position due to the repelling terms in the CA mechanism. Therefore, this research considers two scenarios artificially generated in order to include the aforementioned situations, and the GA parameters for the optimization and simulation parameters for the robots are listed in Tables \ref{tab:ga_parameters} and \ref{tab:parameters_train}.

\begin{table}[htbp]
    \centering
    \caption{GA parameters and algorithms for training}
    \label{tab:ga_parameters}
    \begin{tabular}{cc|c}
        \hline
        \multicolumn{2}{c|}{Parameters} & Values \\ \hline
        \multicolumn{2}{c|}{Number of generations} & 100 \\
        \multicolumn{2}{c|}{Population size} & 64 \\
        \multicolumn{2}{c|}{Stall number of generations} & 20 \\
        \multicolumn{2}{c|}{Elitism ratio} & 0.2\\
        \multicolumn{2}{c|}{Selection algorithm} & Tournament selection \\
        \multicolumn{2}{c|}{Crossover algorithm} & Two-points crossover \\
        \multicolumn{2}{c|}{Mutation algorithm} & Adaptive feasible \\
        \hline
        \end{tabular}
\end{table}

\begin{table}[htbp]
    \centering
    \caption{Simulation parameters for training}
    \label{tab:parameters_train}
    \begin{tabular}{ccc|c}
        \hline
        \multicolumn{3}{c|}{Parameters} & Values \\ \hline
        \multirow{4}{*}{\shortstack{Robots'\\initial \\ \& \\target\\position}} & \multirow{2}{*}{Scenario 1} & Initial & [9, 11]$^T$, [11, 11]$^T$, [11, 9]$^T$, \& [9, 9]$^T$ (m)\\
        & & Target & [81, 81]$^T$, [81, 79]$^T$, [79, 79]$^T$, \& [79, 81]$^T$ (m)\\
        & \multirow{2}{*}{Scenario 2} & Initial & [81, 11]$^T$, [81, 9]$^T$, [79, 9]$^T$, \& [79, 11]$^T$ (m)\\
        & & Target & [25, 81.41]$^T$, [26.41, 80]$^T$, [25, 78.59]$^T$, \& [23.59, 80]$^T$ (m)\\
        \multicolumn{3}{c|}{Target tolerance range} & 0.2 (m)\\
        \multicolumn{3}{c|}{Collision threshold} & 1 (m)\\
        \hline
        \end{tabular}
\end{table}

The location of each robot is posed at each vertex around the square-shaped object with a width of 2 m, and 
it considers a collision if the relative distance between the robot and the nearest obstacle is less than or equal to 1 m that is called collision threshold. In addition, if all robots are within the tolerance range from the target position, it regards the mission completes.

\begin{figure}[htbp]
    \centering
    \begin{subfigure}[b]{0.49\columnwidth}
        \centering
        \includegraphics[width=\columnwidth]{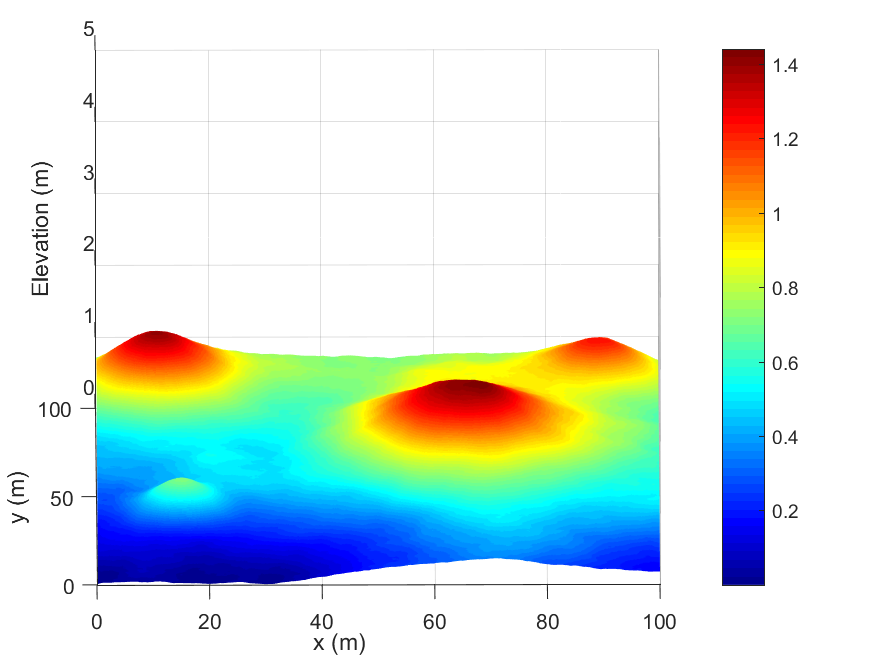}
        \caption{Elevation map}
        \label{fig:test_env_3d}
    \end{subfigure}
    \begin{subfigure}[b]{0.50\columnwidth}
        \centering
        \includegraphics[width=\columnwidth]{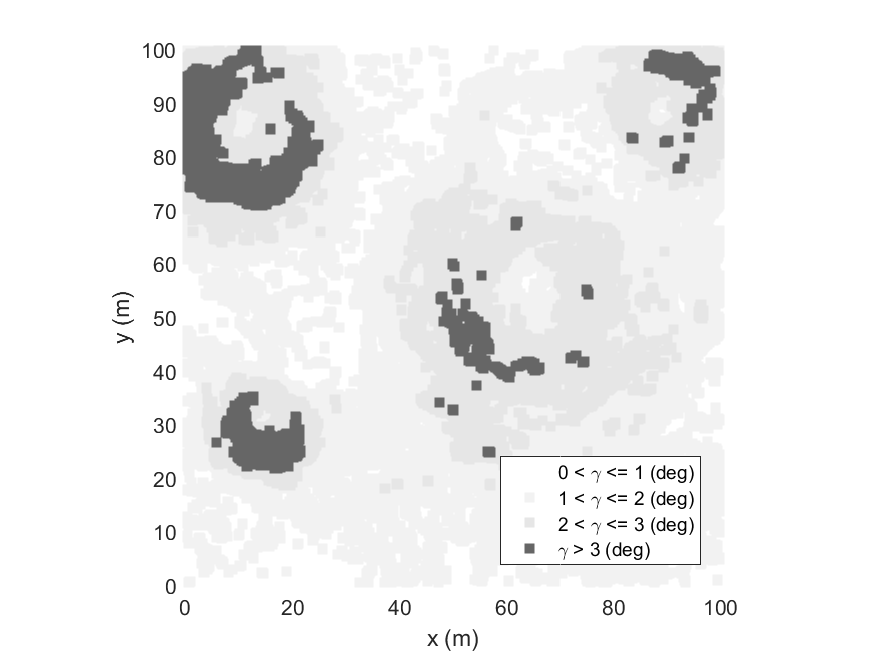}
        \caption{Terrain traversability map}
        \label{fig:test_env_2d}
    \end{subfigure}
    \caption{Testing environment}
    \label{fig:env_testing}
\end{figure}

For the testing environment, it considers the {\it Fractional Brownian surface} as an alternative option instead of the DEM because it can model rough surfaces and complex shapes of nature. The TTA is performed for the environment to convert into the traversability map, and the non-traversable area is assumed as the slope that is greater than 3 deg in the TTA. That is, this research assumes the MRS can make the object with stable attitude for up to 3 deg slope changes. The 3-D elevation map and the converted 2-D terrain traversability map are displayed in Figure \ref{fig:env_testing}. The elevation information of the map is highlighted as a colored surface in Figure \ref{fig:test_env_3d}, and Figure \ref{fig:test_env_2d} shows the slope information at each position. The dark grey regions are non-traversable areas that are considered obstacles because of the slope of more than 3 deg. With this traversability map, the trained FIS models are tested with the parameters listed in Table \ref{tab:parameters_testing}.

\begin{table}[htbp]
    \centering
    \caption{Simulation parameters for testing}
    \label{tab:parameters_testing}
    \begin{tabular}{ccc|c}
        \hline
        \multicolumn{3}{c|}{Parameters} & Values \\ \hline
        \multirow{6}{*}{\shortstack{Robots'\\initial \\ \& \\target\\position}} & \multirow{2}{*}{Scenario 1} & Initial & [15, 11.41]$^T$, [16.41, 10]$^T$, [15, 8.59]$^T$, \& [13.59, 10]$^T$ (m)\\
        & & Target & [31, 81]$^T$, [31, 79]$^T$, [29, 79]$^T$, \& [29, 81]$^T$ (m)\\
        & \multirow{2}{*}{Scenario 2} & Initial & [71, 91]$^T$, [71, 89]$^T$, [69, 89]$^T$, \& [69, 91]$^T$ (m)\\
        & & Target & [14, 6]$^T$, [16, 6]$^T$, [16, 4]$^T$, \& [14, 4]$^T$ (m)\\
        & \multirow{2}{*}{Scenario 3} & Initial & [11, 41]$^T$, [11, 39]$^T$, [9, 39]$^T$, \& [9, 41]$^T$ (m)\\
        & & Target & [90, 31.41]$^T$, [91.41, 30]$^T$, [90, 28.59]$^T$, \& [88.59, 30]$^T$ (m)\\
        \hline
        \end{tabular}
\end{table}

\subsection{Simulation results}

\begin{figure}[htbp]
    \centering
    \begin{subfigure}[b]{0.41\columnwidth}
        \centering
        \includegraphics[width=\columnwidth]{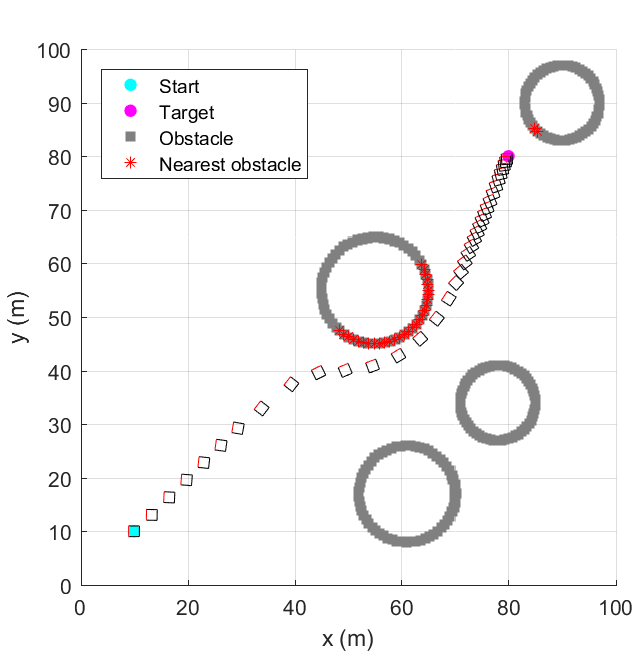}
        \caption{2-D trajectory}
        \label{fig:train1_traj}
    \end{subfigure}
    \begin{subfigure}[b]{0.58\columnwidth}
        \centering
        \includegraphics[width=\columnwidth]{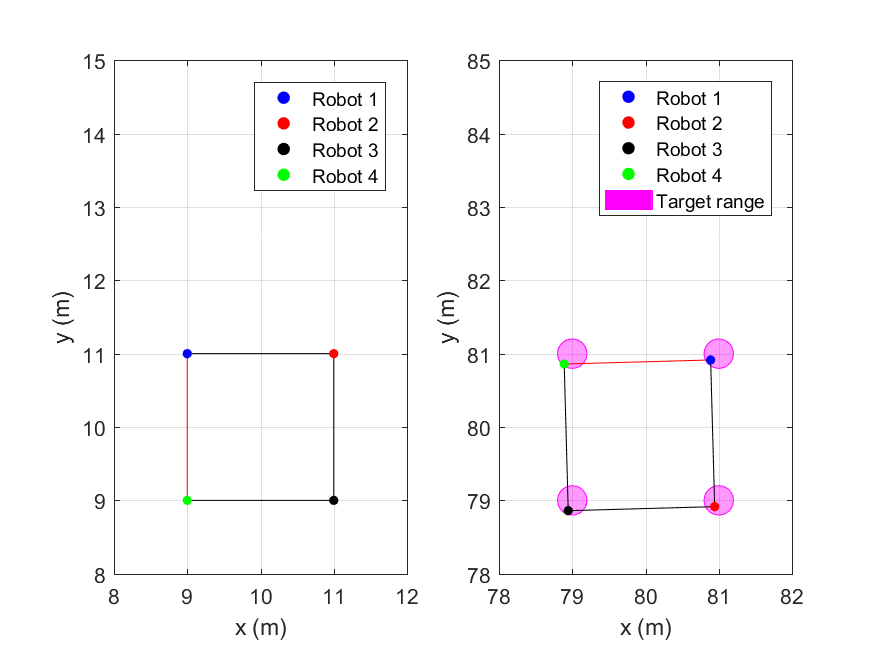}
        \caption{Robots' initial and final position}
        \label{fig:train1_robot_pos}
    \end{subfigure}
    \caption{Trajectory results for the training scenario 1}
    \label{fig:train1_1}
\end{figure}

\begin{figure}[htbp]
    \centering
    \begin{subfigure}[b]{0.5\columnwidth}
        \centering
        \includegraphics[width=0.94\columnwidth]{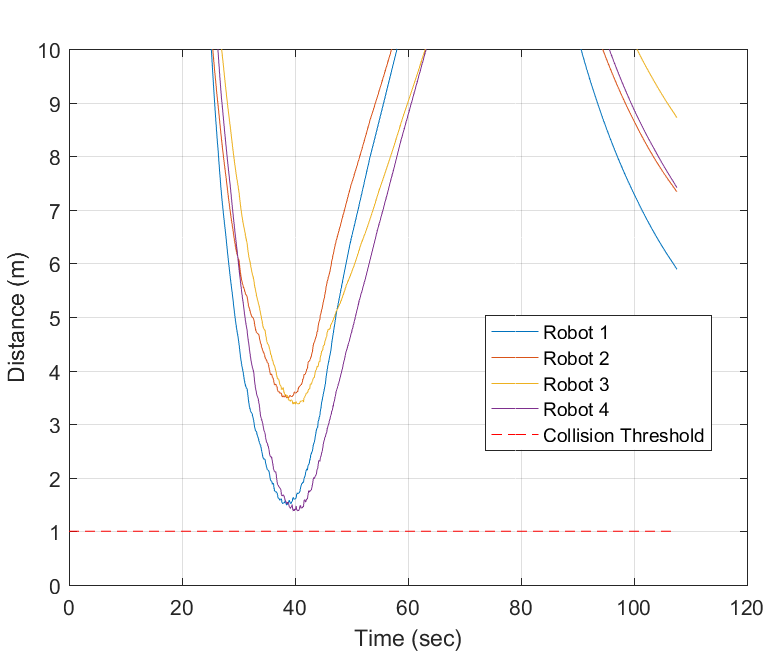}
        \caption{Distance between the robot and the nearest obstacle}
        \label{fig:train1_dist}
    \end{subfigure}
    \begin{subfigure}[b]{0.48\columnwidth}
        \centering
        \includegraphics[width=\columnwidth]{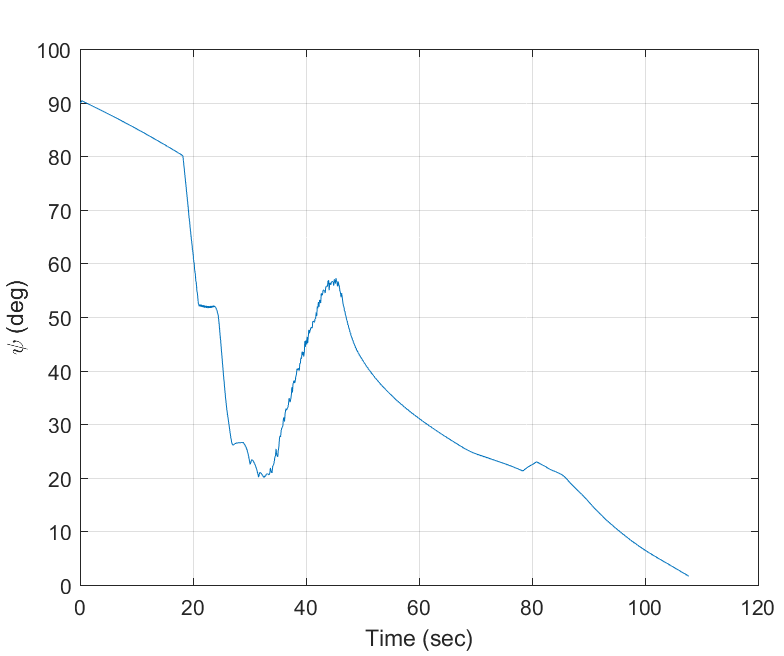}
        \caption{Object's attitude information}
        \label{fig:train1_att}
    \end{subfigure}
    \caption{Time history results for the training scenario 1}
    \label{fig:train1_2}
\end{figure}

\begin{figure}[htbp]
    \centering
    \begin{subfigure}[b]{0.41\columnwidth}
        \centering
        \includegraphics[width=\columnwidth]{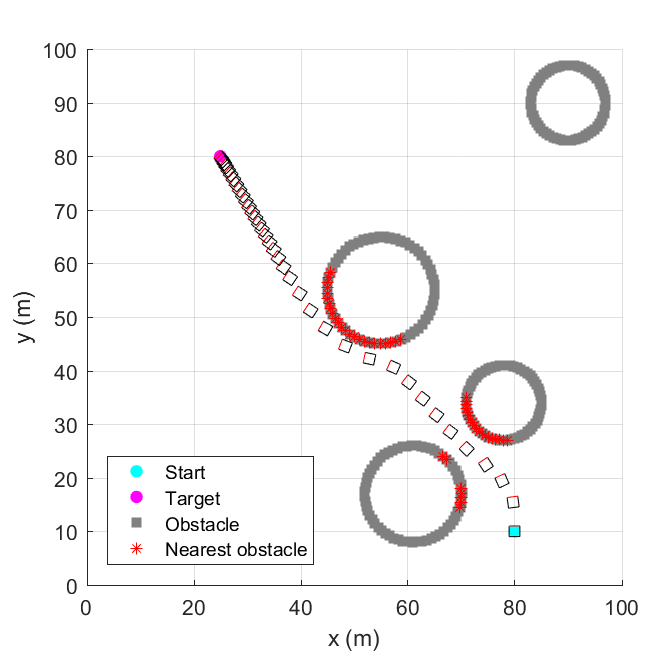}
        \caption{2-D trajectory}
        \label{fig:train2_traj}
    \end{subfigure}
    \begin{subfigure}[b]{0.58\columnwidth}
        \centering
        \includegraphics[width=\columnwidth]{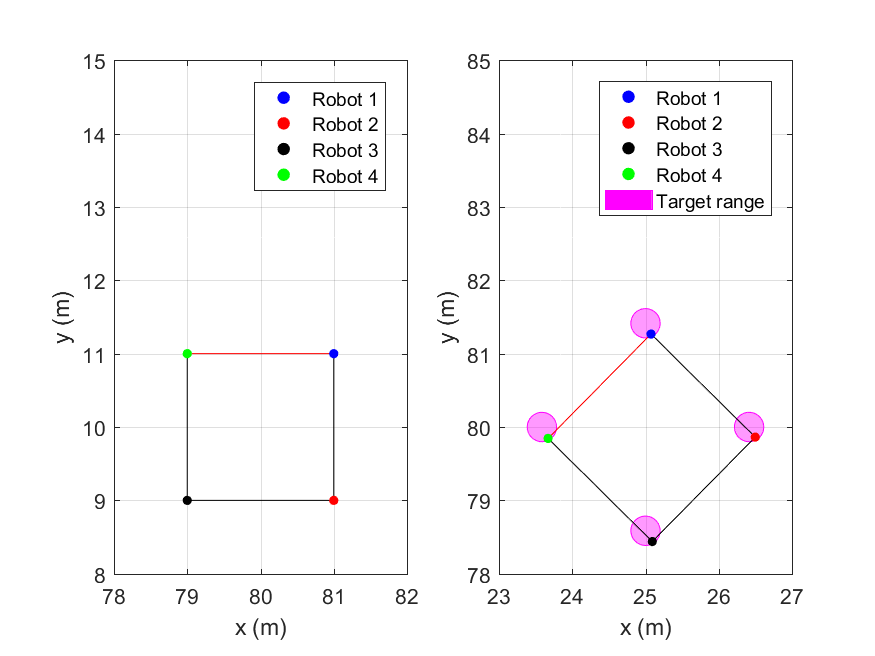}
        \caption{Robots' initial and final position}
        \label{fig:train2_robot_pos}
    \end{subfigure}
    \caption{Trajectory results for the training scenario 2}
    \label{fig:train2_1}
\end{figure}

\begin{figure}[htbp]
    \centering
    \begin{subfigure}[b]{0.5\columnwidth}
        \centering
        \includegraphics[width=0.94\columnwidth]{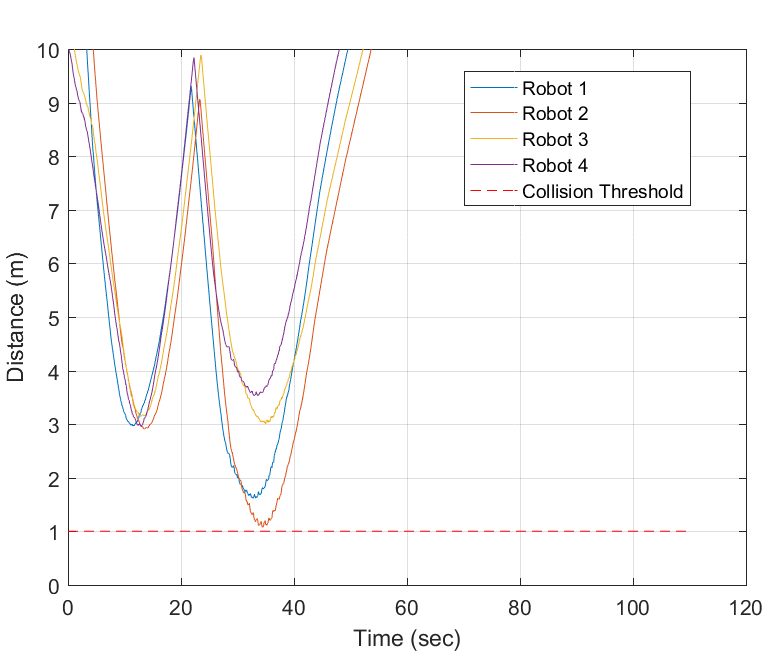}
        \caption{Distance between the robot and the nearest obstacle}
        \label{fig:train2_dist}
    \end{subfigure}
    \begin{subfigure}[b]{0.48\columnwidth}
        \centering
        \includegraphics[width=\columnwidth]{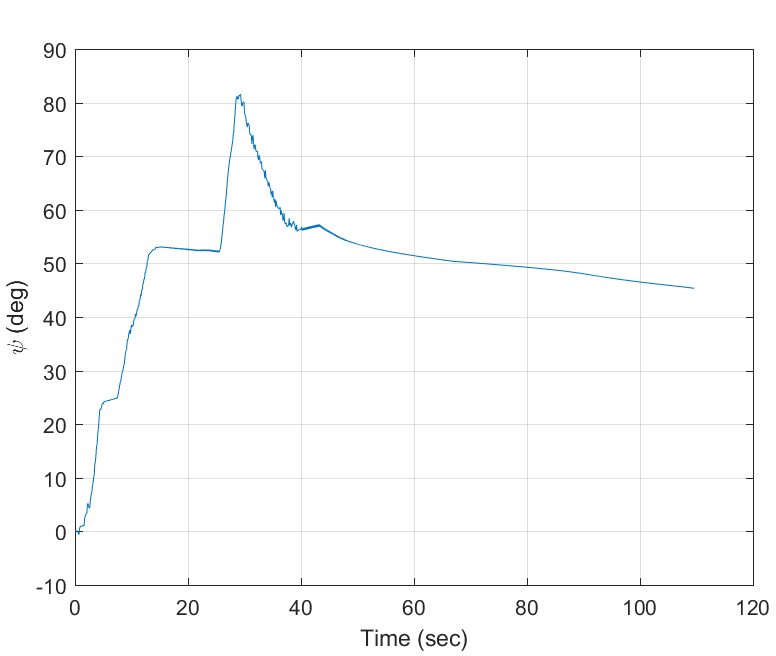}
        \caption{Object's attitude information}
        \label{fig:train2_att}
    \end{subfigure}
    \caption{Time history results for the training scenario 2}
    \label{fig:train2_2}
\end{figure}

With the parameters in Tables \ref{tab:ga_parameters} and \ref{tab:parameters_train}, the training process using the GA is performed, and the training results are displayed in Figures \ref{fig:train1_1} - \ref{fig:train2_2}. Scenario 1 shown in Figures \ref{fig:train1_1} and \ref{fig:train1_2} considers the situations for the local minima and the target that is close to an obstacle, and Scenario 2 shown in Figures \ref{fig:train2_1} and \ref{fig:train2_2} considers a cluttered environment with multiple obstacles. 



Figures \ref{fig:train1_traj} and \ref{fig:train2_traj} show the trajectories of the MRS that reaches the target position without collision, and the total path length of the MRS for each scenario is computed as 419.03 m and 369.23 m, respectively. Figures \ref{fig:train1_robot_pos} and \ref{fig:train2_robot_pos} show the MRS's initial and final position and the object's orientation change, and one can see that the MRS locates within the target tolerance range. Here, the edge between Robot 1 and Robot 4 is highlighted as a red-colored line to easily describe the orientation change of the object in the 2-D trajectories. In addition, figures \ref{fig:train1_dist} and \ref{fig:train2_dist} show the relative distance between each robot and the nearest obstacle, and no collision occurs for both scenarios since the MRS is more than 1 m (collision threshold) away from the obstacle while moving. For the object to be transported, 
the orientation information of the object over time is displayed in Figures \ref{fig:train1_att} and \ref{fig:train2_att}. To achieve the shortest path for the MRS, it is expected that the object's orientation is linearly changed. Here, trend of the orientation changes for two scenarios is roughly observed in linear fashion, but some fluctuations are happened while avoiding obstacles. After passing by the obstacles, the MRS adjusts its velocity and finally meets the target conditions.
 

\begin{figure}[htbp]
    \centering
    \begin{subfigure}[b]{0.43\columnwidth}
        \centering
        \includegraphics[width=\columnwidth]{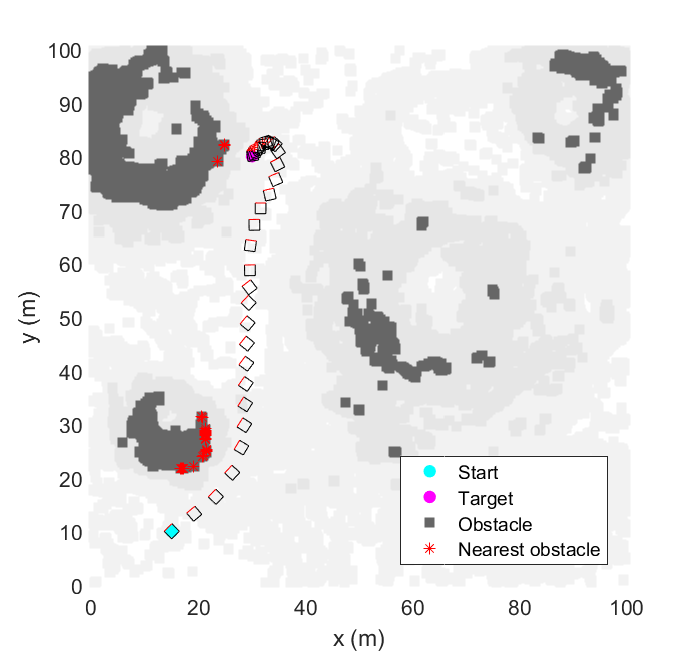}
        \caption{2-D trajectory}
        \label{fig:test1_traj}
    \end{subfigure}
    \begin{subfigure}[b]{0.56\columnwidth}
        \centering
        \includegraphics[width=\columnwidth]{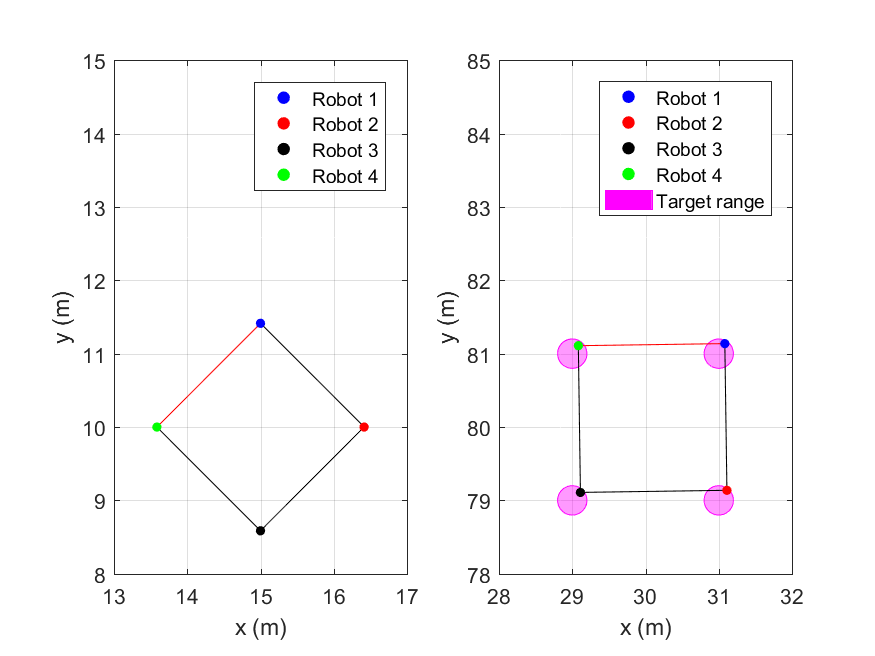}
        \caption{Robots' initial and final position}
        \label{fig:test1_robot_pos}
    \end{subfigure}
    \caption{Trajectory results for the testing scenario 1}
    \label{fig:test1_1}
\end{figure}

\begin{figure}[htbp]
    \centering
    \begin{subfigure}[b]{0.5\columnwidth}
        \centering
        \includegraphics[width=0.94\columnwidth]{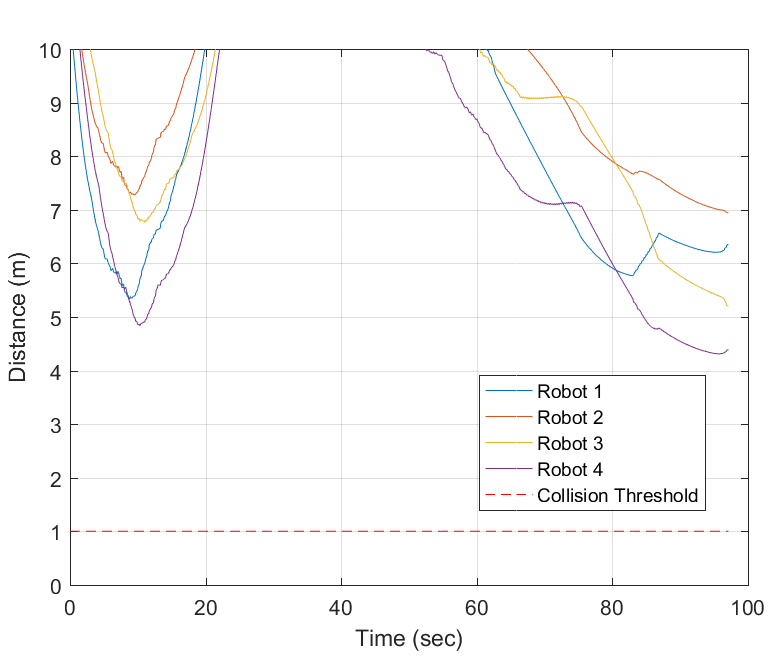}
        \caption{Distance between the robot and the nearest obstacle}
        \label{fig:test1_dist}
    \end{subfigure}
    \begin{subfigure}[b]{0.48\columnwidth}
        \centering
        \includegraphics[width=\columnwidth]{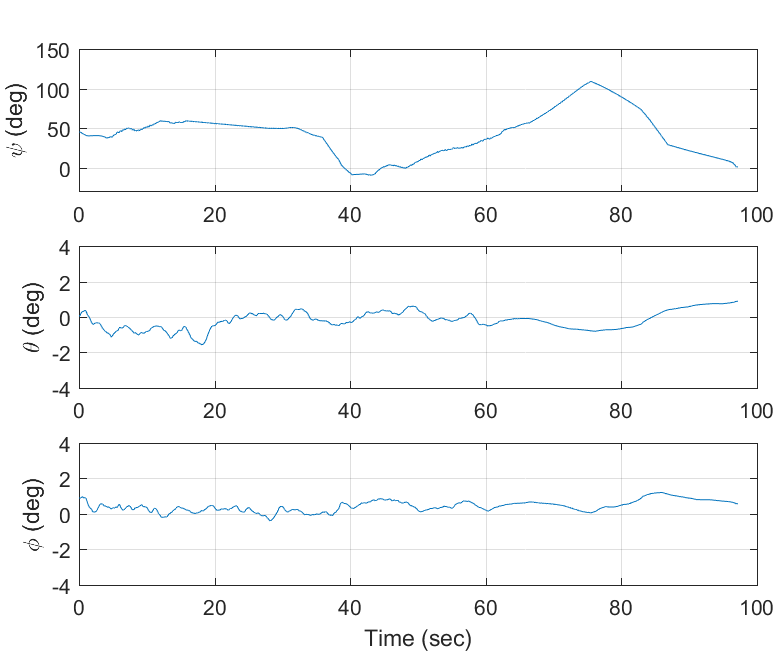}
        \caption{Object's attitude information}
        \label{fig:test1_att}
    \end{subfigure}
    \caption{Time history results for the testing scenario 1}
    \label{fig:test1_2}
\end{figure}


\begin{figure}[htbp]
    \centering
    \begin{subfigure}[b]{0.43\columnwidth}
        \centering
        \includegraphics[width=\columnwidth]{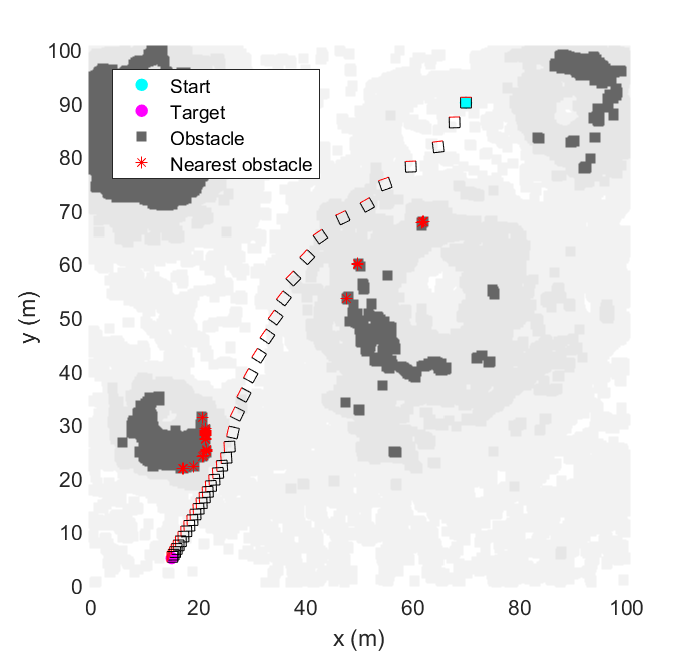}
        \caption{2-D trajectory}
        \label{fig:test2_traj}
    \end{subfigure}
    \begin{subfigure}[b]{0.56\columnwidth}
        \centering
        \includegraphics[width=\columnwidth]{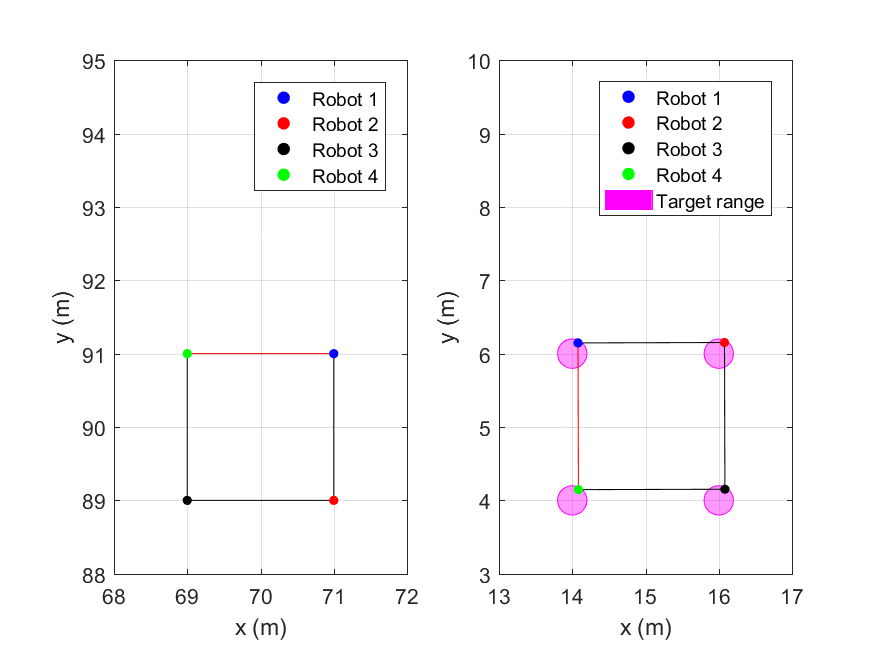}
        \caption{Robots' initial and final position}
        \label{fig:test2_robot_pos}
    \end{subfigure}
    \caption{Trajectory results for the testing scenario 2}
    \label{fig:test2_1}
\end{figure}

\begin{figure}[htbp]
    \centering
    \begin{subfigure}[b]{0.5\columnwidth}
        \centering
        \includegraphics[width=0.94\columnwidth]{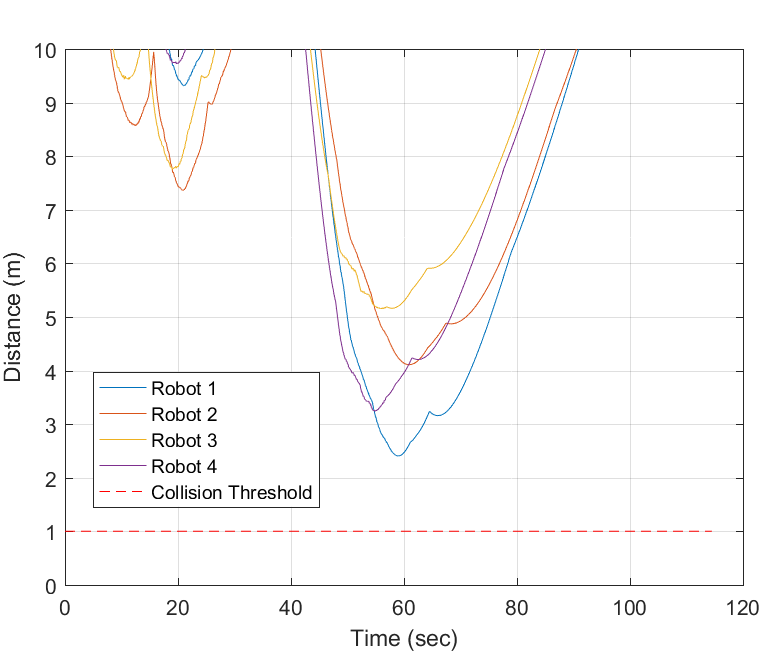}
        \caption{Distance between the robot and the nearest obstacle}
        \label{fig:test2_dist}
    \end{subfigure}
    \begin{subfigure}[b]{0.48\columnwidth}
        \centering
        \includegraphics[width=\columnwidth]{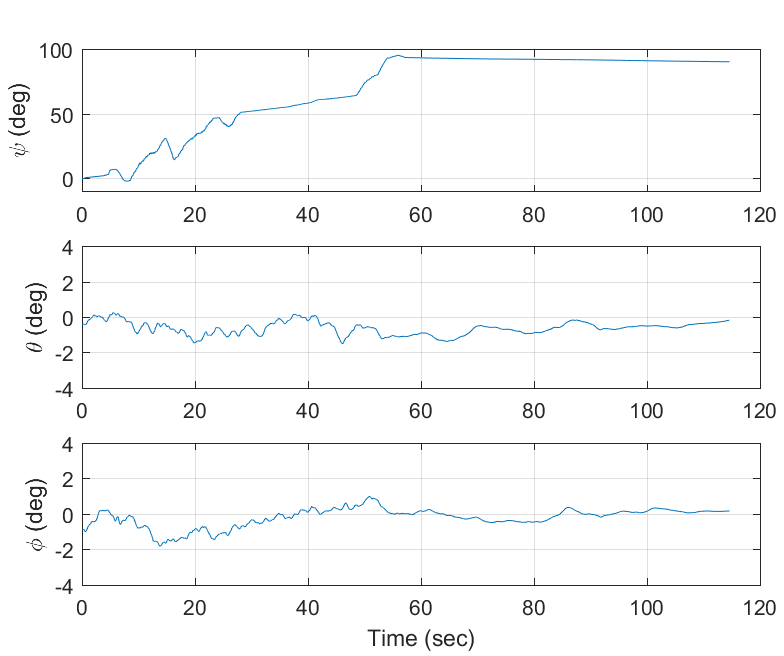}
        \caption{Object's attitude information}
        \label{fig:test2_att}
    \end{subfigure}
    \caption{Time history results for the testing scenario 2}
    \label{fig:test2_2}
\end{figure}


\begin{figure}[htbp]
    \centering
    \begin{subfigure}[b]{0.43\columnwidth}
        \centering
        \includegraphics[width=\columnwidth]{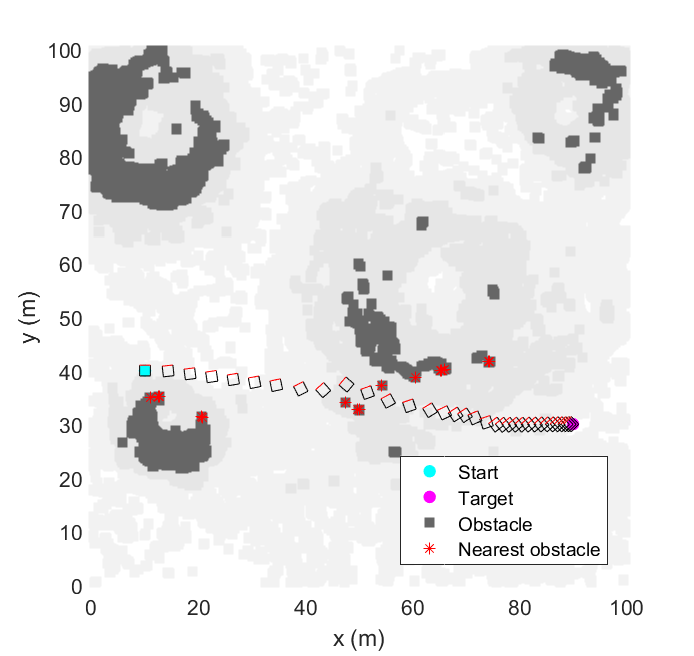}
        \caption{2-D trajectory}
        \label{fig:test3_traj}
    \end{subfigure}
    \begin{subfigure}[b]{0.56\columnwidth}
        \centering
        \includegraphics[width=\columnwidth]{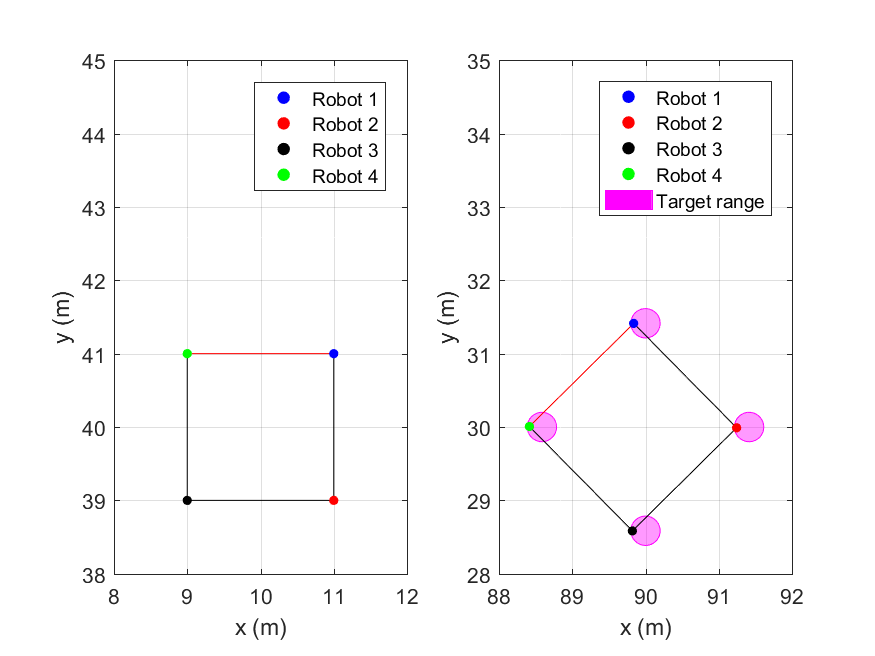}
        \caption{Robots' initial and final position}
        \label{fig:test3_robot_pos}
    \end{subfigure}
    \caption{Trajectory results for the testing scenario 3}
    \label{fig:test3_1}
\end{figure}

\begin{figure}[htbp]
    \centering
    \begin{subfigure}[b]{0.5\columnwidth}
        \centering
        \includegraphics[width=0.94\columnwidth]{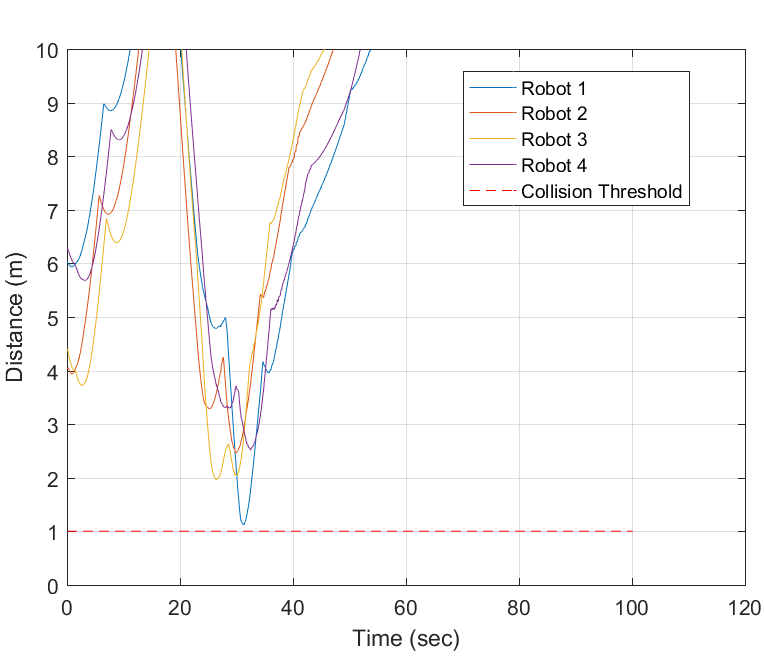}
        \caption{Distance between the robot and the nearest obstacle}
        \label{fig:test3_dist}
    \end{subfigure}
    \begin{subfigure}[b]{0.48\columnwidth}
        \centering
        \includegraphics[width=\columnwidth]{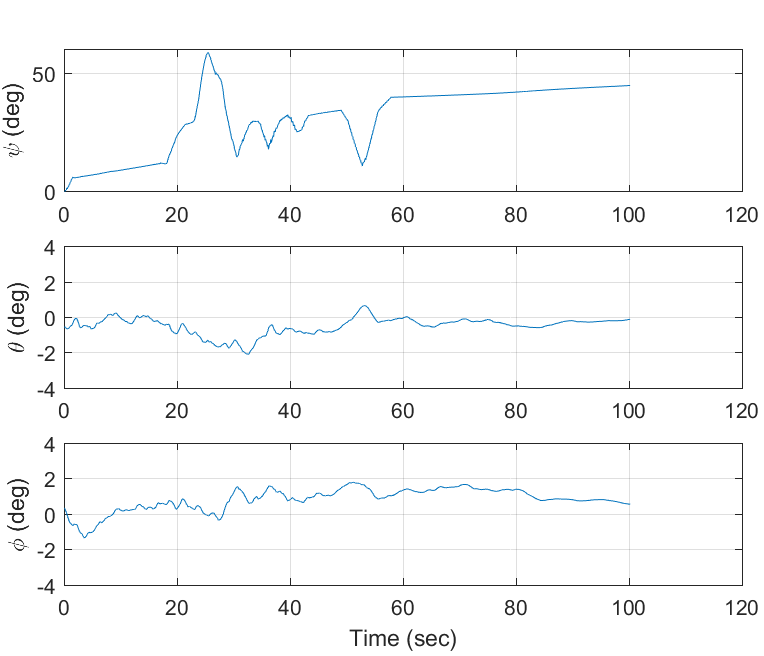}
        \caption{Object's attitude information}
        \label{fig:test3_att}
    \end{subfigure}
    \caption{Time history results for the testing scenario 3}
    \label{fig:test3_2}
\end{figure}

Testing conditions listed in Table \ref{tab:parameters_testing} are applied to trained FIS models in the environment that is the traversability map shown in Figure \ref{fig:test_env_2d}, and the results are displayed in Figures \ref{fig:test1_1} - \ref{fig:test3_2}. The MRS in all scenarios successfully arrives at the target position with no collision as shown in Figures \ref{fig:test1_traj}, \ref{fig:test2_traj}, and \ref{fig:test3_traj}, and the total path length of the robots in each scenario is calculated as 351.84 m, 419.09 m, and 328.59 m, respectively. Figures \ref{fig:test1_robot_pos}, \ref{fig:test2_robot_pos}, and \ref{fig:test3_robot_pos} show the initial and final position of the robots, and it is shown that each robot is located in the target tolerance range at the final time. From the time history of the relative distance for the MRS as shown in Figures \ref{fig:test1_dist}, \ref{fig:test2_dist}, and \ref{fig:test3_dist}, it is observed that all robots avoid obstacles well without collision. In addition, Figures \ref{fig:test1_att}, \ref{fig:test2_att}, and \ref{fig:test3_att} show the object's attitude information that is geometrically calculated by using the robots' $x$ and $y$ position and the elevation information corresponding to the robots' position. As shown in the figures, the roll ($\phi$) and pitch ($\theta$) angles are observed within 3 deg, and it means that the MRS can stably transport the object to the target position. Similar to the training scenarios, the approximate trend of the yaw angle change seems linear, and some fluctuations are observed near obstacles. However, in the case of Scenario 1, the yaw angle of the object reaches about zero angles around 40 sec, and it changes again. In fact, after passing by the first group of obstacles in the bottom-left in Figure \ref{fig:test1_traj}, the MRS adjusts its velocity between 30 and 40 sec and then goes toward the target position directly. However, since obstacles are close to the target position, the magnitude of repelling velocity (${\bf v}_{i,o}$) increases after 40 sec. For this reason, the MRS does not go to the target position directly, but it reaches the target position although the MRS detours. In both training and testing results, the smallest relative distance between the robots and the nearest obstacle is observed in Figures \ref{fig:train2_2} and \ref{fig:test3_2}, and this represents there is a high collision risk in the cluttered environment with multiple obstacles.

\section{Conclusion}
This paper proposes a decentralized multi-robot system (MRS) to perform a collaborative object transportation task in an unstructured environment via the genetic fuzzy system (GFS). The training process is performed in the artificially generated environment with three situations that include the local minima, the target that is close to the obstacle, and the presence of multiple obstacles. With the trained FIS models, another environment that is generated by the Fractional Brownian surface is considered as testing scenarios, and from the train traversability analysis, the given map is transformed into a traversability map that is composed of traversable and non-traversable areas assumed as obstacles. The testing results show that the MRS for all scenarios successfully transport the object to the target position without collision. In addition, this proves that the GFS framework is a good approach for the decentralized MRS to achieve a common task. As future work, the equations of motion of the robots and realistic environment, such as the digital elevation model of Moon or Mars, will be considered, and the consumed energy for each robot will be added to the cost to manage the energy.

\bibliographystyle{AAS_publication}
\bibliography{references}

\end{document}